\pdfoutput=1

\documentclass[11pt]{article}

\usepackage{emnlp2021}

\usepackage{times}
\usepackage{latexsym}
\usepackage{todonotes}
\usepackage{dsfont}
\usepackage{lipsum}
\usepackage{tabularx}
\usepackage{hyperref}
\usepackage{adjustbox}
\usepackage{graphics}

\usepackage[T1]{fontenc}

\usepackage[utf8]{inputenc}

\usepackage{microtype}

\usepackage[acronym]{glossaries}

\PassOptionsToPackage{normalem}{ulem}
\usepackage{ulem}
\providecolor{added}{rgb}{0,0,1}
\providecolor{deleted}{rgb}{1,0,0}
\newcommand{\added}[1]{{#1}}
\newcommand{\deleted}[1]{{}}

\title{Beyond Accuracy: A Consolidated Tool for Visual Question Answering Benchmarking}

\makeatletter
\newcommand{\printfnsymbol}[1]{%
  \textsuperscript{\@fnsymbol{#1}}%
}
\makeatother

\author{Dirk Väth* \and Pascal Tilli* \and Ngoc Thang Vu \\
        University of Stuttgart \\ Germany \\ 
        \{dirk.vaeth, pascal.tilli, thang.vu\}@ims.uni-stuttgart.de}

\begin{document}

\newacronym{ml}{ML}{Machine Learning}
\newacronym{dl}{DL}{Deep Learning}
\newacronym{vqa}{VQA}{Visual Question Answering}
\newacronym{sear}{SEAR}{Semantically Equivalent Adversarial Rules}
\newacronym{butd}{BUTD}{Bottom-up and Top-down Attention }
\newacronym{mfh}{MFH}{Generalized Multimodal Factorized High-Order Pooling}
\newacronym{ban}{BAN}{Bilinear Attention Networks}
\newacronym{mfb}{MFB}{Multi-modal Factorized Bilinear Pooling}
\newacronym{mcan}{MCAN}{Deep Modular Co-Attention Networks}
\newacronym{mmnasnet}{MMNASNET}{Deep Multimodal Neural Architecture Search}
\newacronym{pos}{POS}{Part-of-Speech}
\newacronym{glove}{GloVe}{GloVe}
\newacronym{ft}{FT}{FastText}
\newacronym{bert}{BERT}{BERT}

\newcommand\blfootnote[1]{%
   \begingroup
   \renewcommand\thefootnote{}\footnote{#1}%
   \addtocounter{footnote}{-1}%
   \endgroup
}

\maketitle
\begin{abstract}
On \blfootnote{* Authors contributed equally} the way towards general \gls{vqa} systems that are able to answer arbitrary questions, the need arises for evaluation beyond single-metric leaderboards for specific datasets.
To this end, we propose a browser-based benchmarking tool for researchers and challenge organizers, with an API for easy integration of new models and datasets to keep up with the fast-changing landscape of \gls{vqa}.
Our tool helps test generalization capabilities of models across multiple datasets, evaluating not just accuracy, but also performance in more realistic real-world scenarios such as robustness to input noise.
Additionally, we include metrics that measure biases and uncertainty, to further explain model behavior.
Interactive filtering facilitates discovery of problematic behavior, down to the data sample level.
As proof of concept, we perform a case study on four models.
We find that state-of-the-art \gls{vqa} models are optimized for specific tasks or datasets, but fail to generalize even to other in-domain test sets, for example they cannot recognize text in images.
Our metrics allow us to quantify which image and question embeddings provide most robustness to a model.
All code\footnote{https://github.com/patilli/vqa\_benchmarking} is publicly available.
\end{abstract}

\section{Introduction}

\Gls{vqa} refers to the multi-modal task of answering free-form, natural language questions about images - a task sometimes referred to as a visual Turing test \cite{xu2018fooling}.
\begin{figure}
    \centering
    \includegraphics[width=0.49\textwidth]{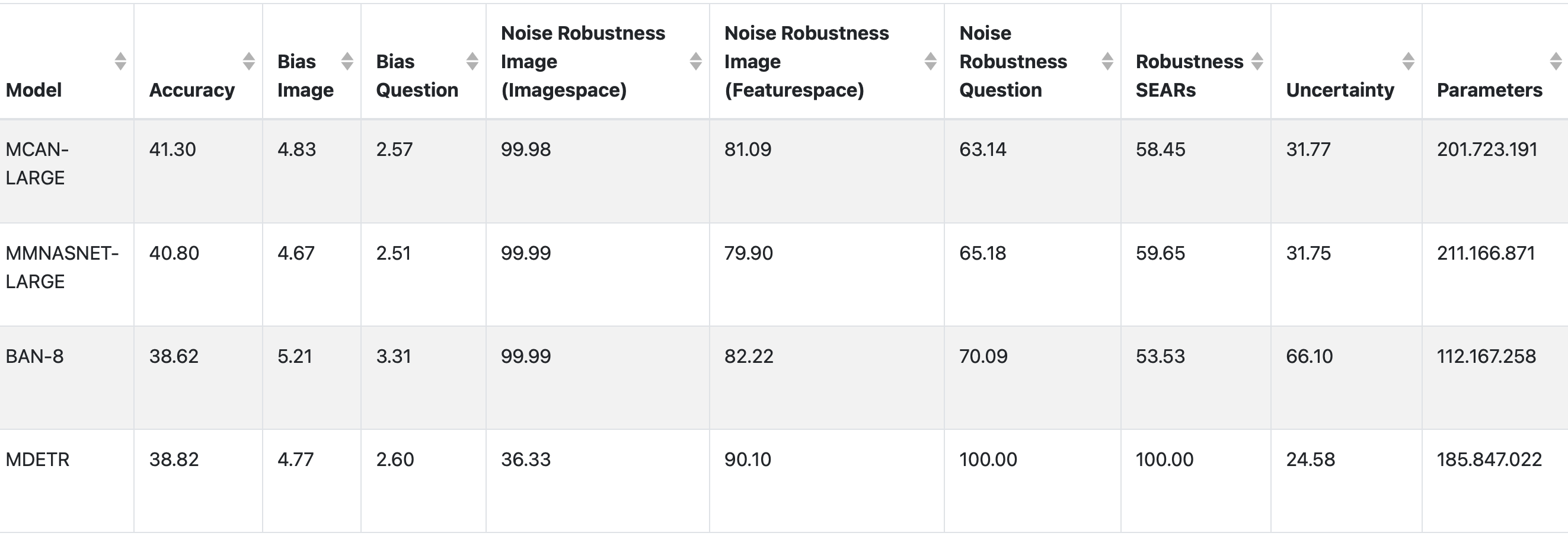}
    \caption{Tool landing page with aggregated metrics for each model (larger version in Appendix A). Statistics can be expanded per model to view performance on each dataset.}
    \label{fig:overview}
\end{figure}
The number and variety of datasets for evaluating such systems has continued to increase over the last years \cite{antol2015vqa, hudson2019gqa, agrawal2018don, kervadec2021roses, johnson2017clevr}.
These datasets aim to test models' abilities with respect to different skills, such as commonsense or external knowledge reasoning, visual reasoning, or reading text in images.
Traditionally, evaluation relies solely on answering accuracy.
However, it is misleading to believe that a single number, like high accuracy on a given benchmark, corresponds to a system's ability to answer arbitrary questions with high quality.
Each dataset contains biases which state-of-the-art deep neural networks are prone to exploit, resulting in higher accuracy scores \cite{goyal2017making, DAS201790, agrawal2016analyzing, jabri2016revisiting}.
Thus, most \gls{vqa} models, if evaluated on multiple benchmarks at all, are re-trained per dataset to achieve higher numbers in specialized leaderboards.
Further shortcomings of current leaderboards include ignoring prediction cost and robustness, as discussed in \cite{ethayarajh2020utility} for NLP.
In \gls{vqa}, we need even more specialized evaluation due to the challenges inherent to open-ended, multi-modal reasoning.

In order to successfully develop \gls{vqa} systems that are able to answer arbitrary questions with human-like performance, we should overcome the previously mentioned shortcomings of current leaderboards  as one of the first essential steps.
To this end, we propose a benchmarking tool, to our knowledge the first of its kind in the \gls{vqa} domain, that goes beyond current leaderboard evaluations. It follows the four following principles:
\paragraph{1. Realism} To better simulate real-world conditions, we test robustness to semantic-preserving input perturbations to images and questions.
\paragraph{2. Generalizability} We include six carefully chosen benchmark datasets, each evaluating different abilities as well as model behavior on changing distributions to test model generalizability.
Additionally, we provide easy python interfaces for adding new benchmarking datasets and state-of-the-art models as they arise. The full tool is released under an open-source license.
\paragraph{3. Explainability} To provide more insight into model behavior and overcome the problem of single-metric comparisons, we measure scores such as biases and uncertainty, in addition to accuracy.
\paragraph{4. Interactivity} Aggregated statistics can be drilled down and filtered, providing interactive exploration of model behavior from a global dataset perspective down to individual data samples.
The above functionalities not only support detailed model comparison, but also facilitate development and debugging of models in the \gls{vqa} domain.

As proof of concept, we integrate several popular and state-of-the-art models from public code repositories and investigate their abilities and weaknesses.
Through our case study, we demonstrate that all of these models fail to generalize, even to other in-domain test sets.
Our metrics quantify the influence of model architecture decisions, which accuracy cannot capture, such as the effect of image and question embeddings on model robustness.

\section{Related Work}
\textbf{VQA Benchmarks} Benchmarks often emphasize certain sub-tasks of the general \gls{vqa} problem.
For example, CLEVR \cite{johnson2017clevr} tests visual reasoning abilities such as shape recognition and spatial relationships between objects, rather than real-world scenarios.
Other approaches change the answer distributions of existing datasets, such as VQA-CP \cite{agrawal2018don} originating from VQA2 \cite{goyal2017making} or GQA-OOD from GQA \cite{kervadec2021roses}.
These changes are intended to mitigate learnable bias. Another approach to mitigating biases was proposed by \citet{hudson2019gqa}, who created a dataset for real world visual reasoning with a tightly controlled answer distribution.
\citet{kervadec2021roses} went one step further by analyzing and changing the test sets to evaluate on rarely occurring question-answer pairs rather than on frequent ones. Finally, \citet{li2021adversarial} proposed an adversarial benchmarking dataset to evaluate system robustness.

\paragraph{Metrics}
For more automated, dataset-level insight, many methods try to analyze single aspects of \gls{vqa} models. For example, \citet{halbe2020exploring} use feature attribution to assess the influence of individual question words on model predictions.
\citet{DAS201790} compare human attention to machine attention to explore whether they focus on the same image regions.
To measure robustness w.r.t. question input, \citet{huang2019novel} collect a dataset of semantically relevant questions, and rank them by similarity, feeding the top-3 into the network to observe changes in prediction.

\paragraph{Identifying Biases} 
\citet{agrawal2016analyzing} measured multiple properties: generalization to novel instances as selected by dissimilarity, question understanding based on length and POS-tags, and image understanding by selecting a subset of questions which share an answer but have different  images.
Another approach to analyze bias towards one modality is to train a uni-modal network that  excludes the other modality in the training phase \cite{unimodal_biases}.
However, this requires training one model per modality and cannot be applied easily to all architectures, e.g. to attention mechanisms computed on joint feature spaces.

\paragraph{Robustness and Adversarial Examples}
Adversarial examples originate from image classification, where perturbations barely visible to a human fool the classifier and cause sudden prediction changes \cite{adversarial}.
The same idea was later applied to NLP, where, e.g., appending distracting text to the context in a question answering scenario resulted in F1-score dropping by more than half \cite{jia2017adversarial}.

\paragraph{Benchmarking Tools}
\citet{liu2021explainaboard} propose a leaderboard for NLP tasks to compare model performance.
They differentiate among several NLP tasks and datasets.
All methods are applied post-hoc to analyze the predictions a model outputs.
Other benchmarking tools, for example, focus on runtime comparisons \cite{shi2016benchmarking, liu2018benchmarking}. Our benchmarking tool not only analyzes system outputs, but also modifies input modalities as well as feature spaces and provides metrics beyond just accuracy.

\section{VQA Benchmark Tool}

Our tool facilitates global evaluation of model performance w.r.t general and specific tasks (\textbf{generalizability}), such as real-world images and reading capabilities. 
To simplify integration of future benchmark datasets and models, we provide a well-documented python API.
We measure model-inherent properties, such as biases and uncertainty (\textbf{explainability}) as well as robustness against input perturbations (\textbf{realism}).
Model behavior can be further inspected using interactive diagrams and filtering methods for all metrics, supporting sample-level exploration of suspicious model behavior (\textbf{interactivity}).
All data is collected post-hoc and can be explored in a web application, eliminating the need to re-train existing models.

\subsection{Datasets}
In this section, we describe the datasets supported out-of-the-box.
These serve the principle of benchmarking \textbf{generalizability}, by including real-world scenarios as well as task-specific and even synthetic datasets. 
Where labels are publicly available, we rely on test sets, otherwise on validation sets (marked with ${}^\ast$).
To reduce computational cost and resources (including environmental impact), we limit each dataset to a maximum of $\sim$15,000 randomly drawn samples \added{, which is referred to in the following paragraphs as a sub-set}.

\paragraph{VQA2$^\ast$}
This dataset \cite{goyal2017making} represents a balanced version of the vanilla VQA dataset \cite{antol2015vqa}. 
It is intended to mirror real-world scenarios and used as the de-facto baseline for model comparisons.

\paragraph{GQA}
The GQA dataset \cite{hudson2019gqa}, derived from Visual Genome \cite{krishna2017visual}, is designed to test models' real-word visual reasoning capabilities, in particular, robustness, consistency and semantic understanding of vision and language.
Similar to VQA2, it also provides a balanced version.

\paragraph{GQA-OOD}
According to \citet{kervadec2021roses}, evaluating on rare instead of frequent question-answer pairs is better suited for measuring reasoning abilities.
Hence, they introduce the GQA-OOD dataset as a new split of the original GQA dataset to evaluate out-of-distribution questions with imbalanced distributions.

\paragraph{CLEVR$^\ast$}
CLEVR \cite{johnson2017clevr} is a synthetic dataset, containing images of multiple geometric objects in different colors, materials and arrangements.
It aims to test models' visual reasoning abilities by asking questions that require a model to identify objects based on attributes, presence, count, and spacial relationships.

\paragraph{OK-VQA$^\ast$}
\citet{marino2019ok} introduce a dataset that requires external knowledge to answer its questions, thereby motivating the integration of additional knowledge pools.

\paragraph{Text VQA$^\ast$}
\citet{singh2019towards} consider the problem of understanding text in images,
an important problem to consider in \gls{vqa} benchmarking systems, as one application of \gls{vqa} is intended to aid the visually impaired.

\subsection{Metrics}
\label{sec:tool:metrics}

In addition to the evaluation of accuracy across datasets with different distributions and focuses, we implement metrics such as bias of models towards one modality and uncertainty (\textbf{explainability}), as well as robustness to noise and adversarial questions (\textbf{realism}). \added{All metrics are in range $[0,100]$.}

\paragraph{Accuracy}
Our tool supports multiple ground truth answers with different scores \added{per sample}, providing the flexibility to evaluate for single-answer accuracy as well as e.g. the official VQA2 accuracy measure $acc(a) = min(1, \frac{\#\text{humans}(a)}{3})$ \added{\cite{antol2015vqa}}.

\paragraph{Modality Bias}
\label{sec:tool:mbias}
Here, we refer to a model's focus on one modality over the other.
Given an image of a zoo and the question \emph{``What animals are shown?''}, if we replace this picture with a fruit bowl, we would expect the model to change its prediction.
However, if the prediction stays unaltered, the model's answer cannot depend on the image input.
For each prediction on altered inputs $(i',q)$ or $(i, q')$, we evaluate how many times the answer $a'$ of the replacement pairs is the same as answer $a$ predicted on the original inputs $(i,q)$.
Averaging across $N$ trials yields a Monte-Carlo estimate of the bias towards one modality as $1/N \sum_{q'} \mathds{1}_{f(q,i)=f(q',i)}$.
Heuristics, such as ensuring no overlap between subjects and objects of $q$ and $q'$, help reduce cases where $q'$ would just be a rephrasing of $q$.
\added{High values in modality bias correspond to models ignoring input from one modality for many samples, e.g. a question bias of $100$ indicates a model that completely ignores images.}

\paragraph{Robustness to Noise}
An important consideration when deploying a model in the real world, is its susceptibility to noise.
Noise might be induced naturally by data acquisition methods (\gls{vqa}-setting: camera), for example a color-question should not be affected by subtle tone shifts between two cameras.
On the question side, semantic-preserving input changes can be induced through paraphrases, synonyms or region-dependent spelling.

For measuring robustness to noise in images, we support adding Gaussian-, Poisson-, salt\&pepper- and speckle-noise to the original input image.
We also support adding Gaussian noise in image feature space. To obtain a realistic input range, we calculate the standard deviation from 500 randomly sampled image feature vectors.
After multiple trials, we average how often the prediction on the noisy inputs matches the original prediction.
Applying noise to the original image input tests the robustness of the image feature extractor, which, in many models, is external and thus easy to swap and interesting to compare.
On the other hand, applying noise in feature space tests model robustness towards noisy feature extractors.

Measuring robustness to question noise is done by adding Gaussian noise in embedding space, a reasonable approach under the assumption that similar vectors in embedding space have similar meaning.
Again, multiple trials are performed.

\added{High values in robustness correspond to models unaffected by noise in one modality for many samples, e.g. a question robustness of $100$ indicates a model that never changed its predictions due to noise added in question embedding space. }

\paragraph{Robustness to Adversarial Questions}

\Glspl{sear} alter textual input according to a set of rules, while preserving original semantics \cite{sears}.
For the questions in the \gls{vqa} dataset, the authors come up with the four rules that most affect the predictions in their tests, using a combination of \gls{pos}-Tags and vocabulary entries:

\begin{itemize}
\item \emph{Rule 1} $\text{WP VBZ} \rightarrow \text{WP's}$
\item \emph{Rule 2} $\text{What NOUN} \rightarrow \text{Which NOUN}$
\item \emph{Rule 3} $\text{color} \rightarrow \text{colour}$
\item \emph{Rule 4} $\text{ADV VBZ} \rightarrow \text{ADV's}$
\end{itemize}

\added{High values in robustness against \glspl{sear} correspond to models unaffected by adversarial questions, e.g. a robustness of $100$ indicates a model that never changed its predictions due to the application of any of the above rules. Therefore, higher values are preferable.}

\paragraph{Uncertainty}

To measure model certainty, we leverage the dropout-based Monte-Carlo method \cite{mcdropout}.
Forwarding a sample multiple times with active dropout, the averaged output vector $\frac{1}{N} \sum_{n=1}^{N} f(x)$.

\subsection{Views}
We support inspection of the included metrics at different levels of granularity, from comparisons across multiple datasets to filtering of individual samples (\textbf{interactivity}).
On each level, we supplement the accuracy measure by additional metrics helpful for understanding and debugging \gls{vqa} models (\textbf{explainability}).

\paragraph{Gobal View}
The global view (see figure \ref{fig:overview}) acts as the main entry to our tool.
At a glance, it shows a leaderboard with statistics averaged on all datasets, providing users with an impression of the models' performance and properties across tasks and distributions.
\begin{figure}[htb]
    \centering
    \includegraphics[width=0.49\textwidth]{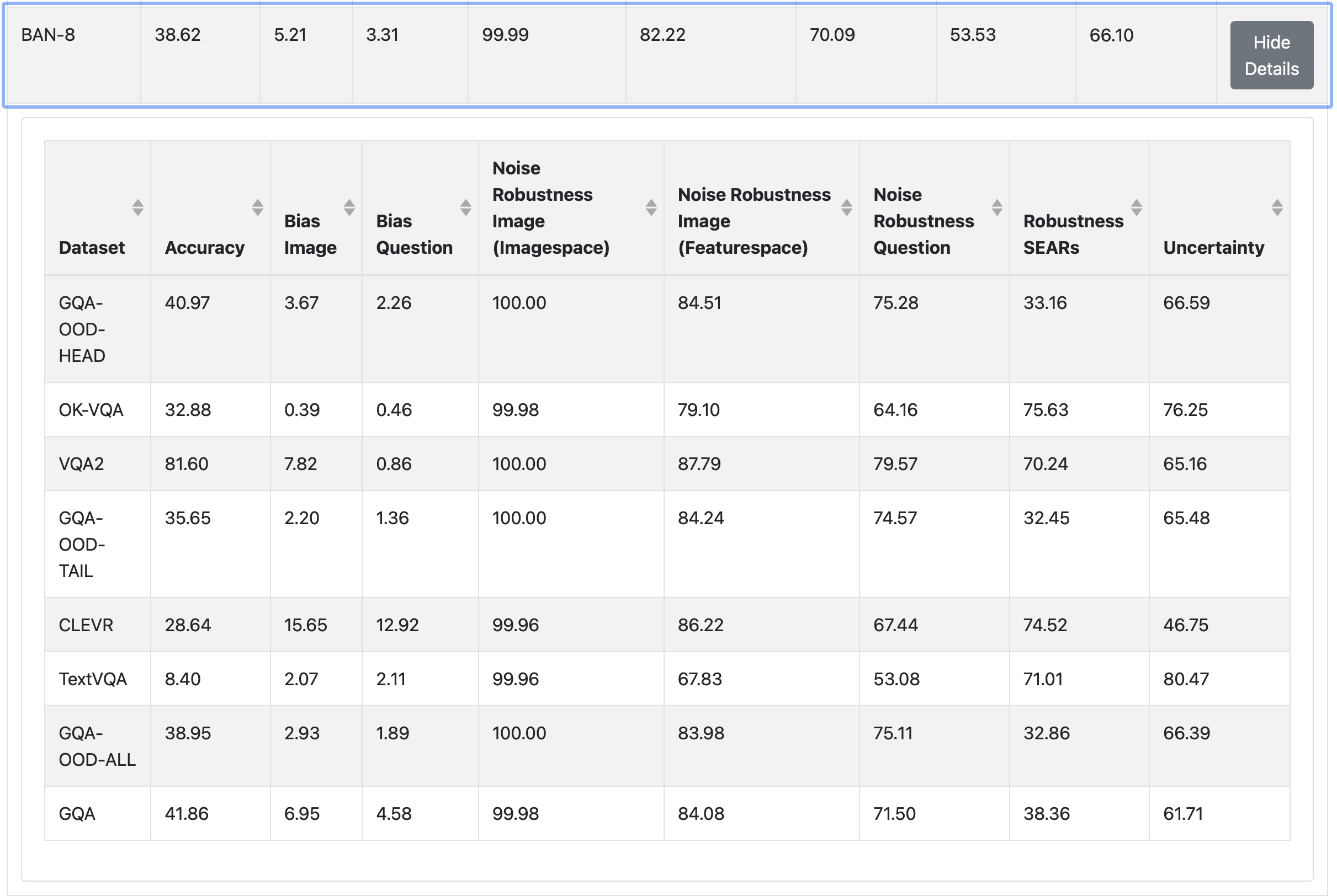}
    \caption{Expanded details on the overview page (larger version in Appendix A).}
    \label{fig:ov_detail_view}
\end{figure}
All columns are sortable to allow easy comparisons between models for each metric.
Each row in the overview table describes a model's average performance and can be expanded to provide additional information on a per-dataset level (see figure \ref{fig:ov_detail_view}).

\paragraph{Metrics View}
\begin{figure}[htb]
    \centering
    \includegraphics[width=0.5\textwidth]{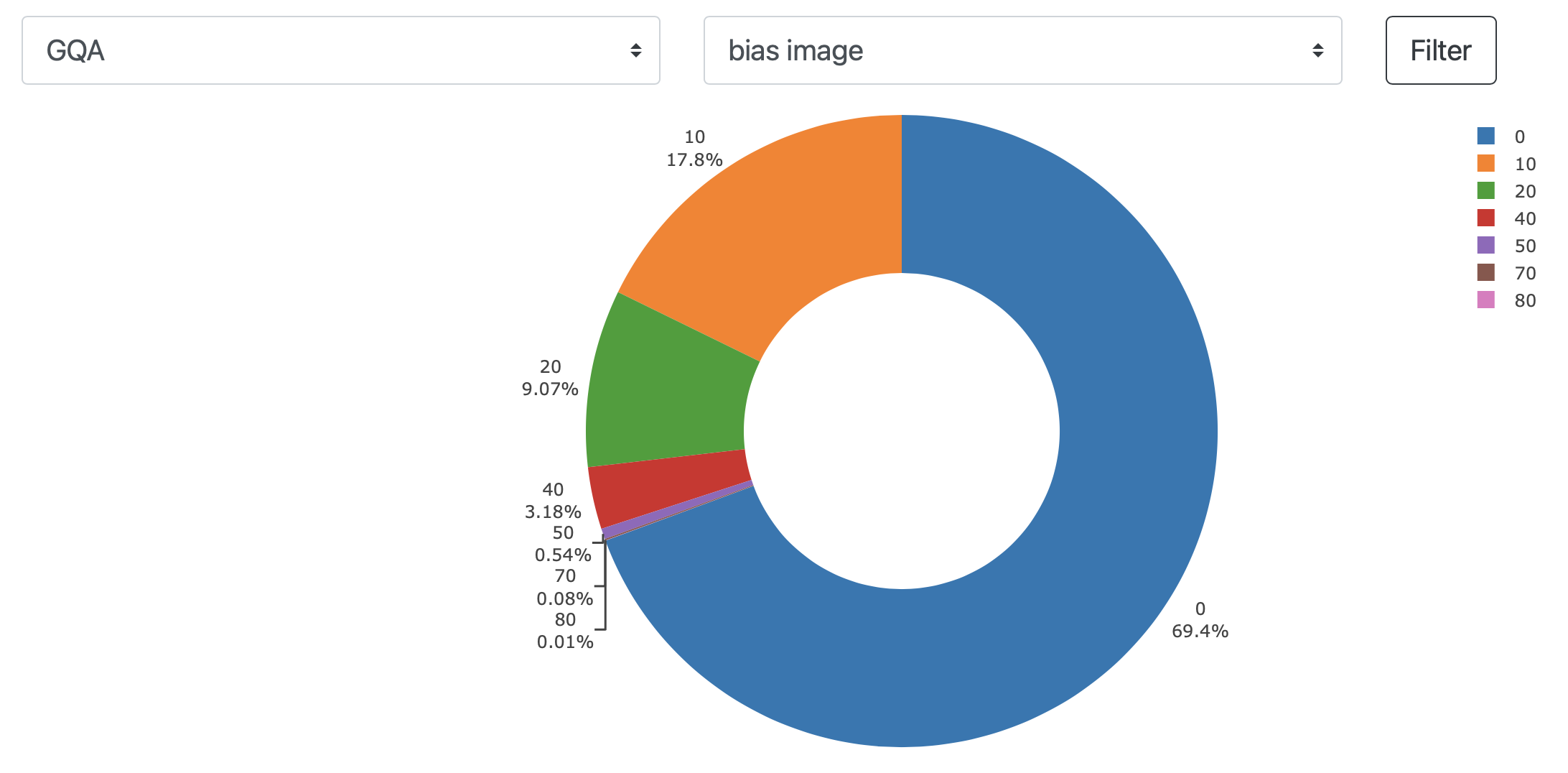}
    \caption{Metric view, showing bias towards images on the GQA dataset for the MDETR model (larger version in Appendix A).}
    \label{fig:metric_view}
\end{figure}
Clicking a model row in the global view navigates to the metrics view, which provides graphs on all metrics and datasets for the selected model in detail (see figure \ref{fig:metric_view}).
Users have the choice to change dataset and metric via selection boxes.
For easy comparison between datasets of different sizes, all values are recorded in percentages of the dataset.

\begin{figure}[htb]
    \centering
    \includegraphics[width=0.49\textwidth]{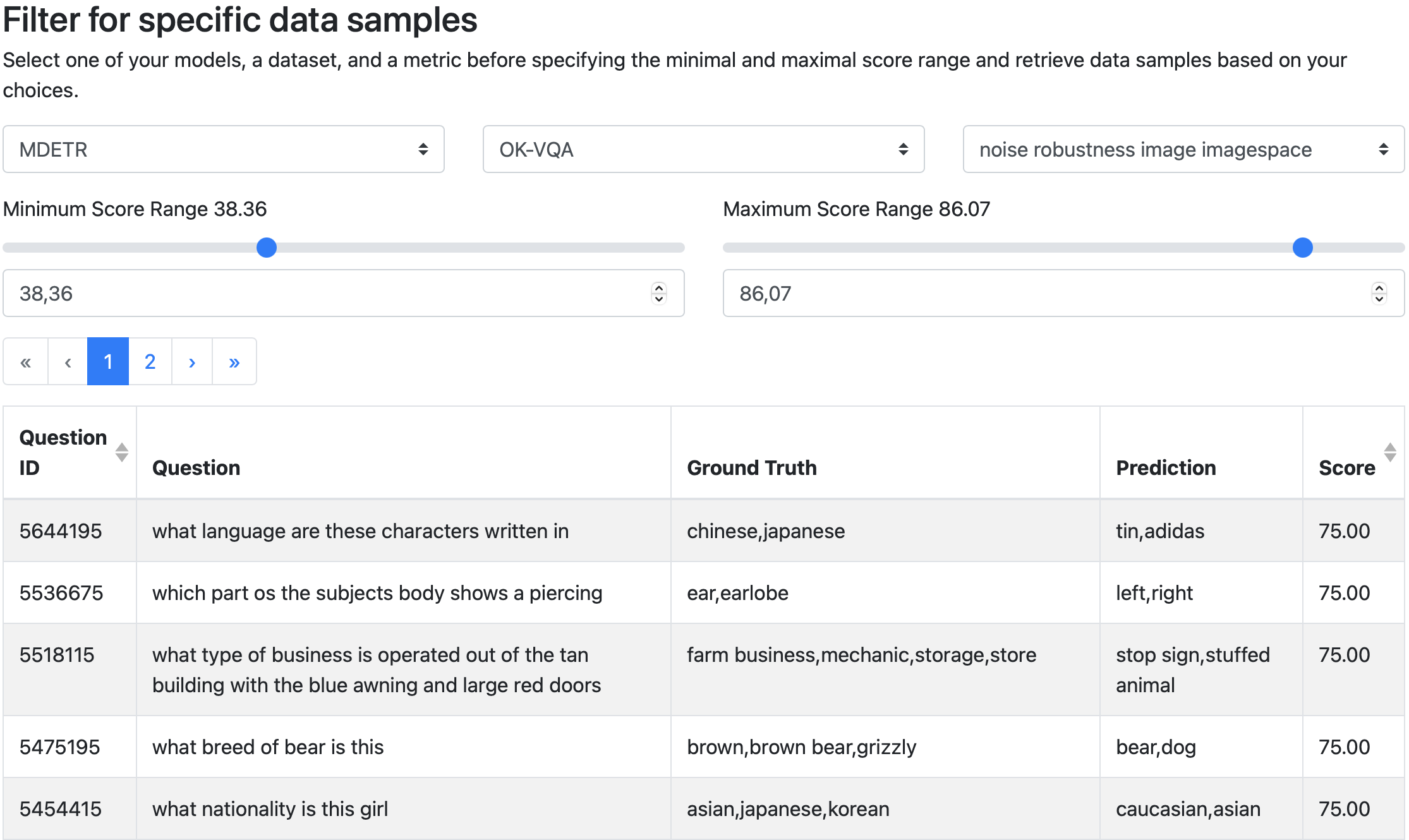} 
    \caption{Filter view. Enables filtering for unexpected model behavior on sample level (larger version in Appendix A).}
    \label{fig:filter_view}
\end{figure}

\paragraph{Filter View}
Our tool supports searching for patterns of suspicious model behavior by providing a filter view (see figure \ref{fig:filter_view}).
Once model, dataset and metric are selected, users are presented a list of all samples within the chosen range.
The range can be adjusted using a slider, which updates the list of matching data samples in real-time.

\begin{figure}[htb]
    \centering
    \includegraphics[width=0.49\textwidth]{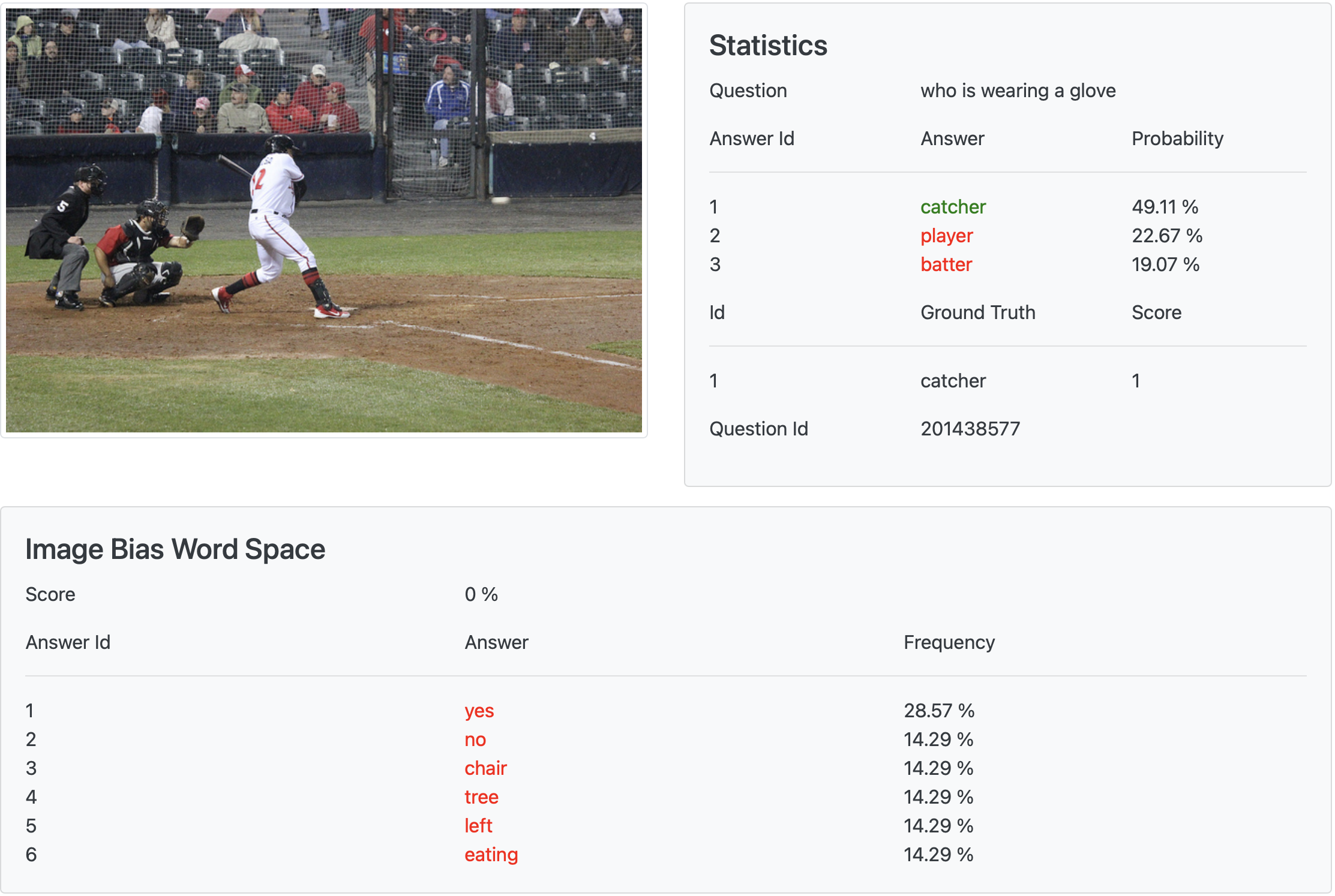}
    \caption{Sample view with \textit{Image Bias Word Space} metric. Similar cards exist for each metric (larger version in Appendix A).}
    \label{fig:sample_view}
\end{figure}

\paragraph{Sample View}
Finally, once the desired range of samples has been filtered, clicking a data sample navigates to the sample view (see figure  \ref{fig:sample_view}).
There, the original input image and question are displayed, along with  ground truth and the model's top-3 predictions.
Additionally, the scores and answers for each single metric are shown.
For example, if applying noise to the image changed the prediction multiple times, we show all the answers that were predicted using those noisy inputs.

\section{Case Study}

As a case study, we explore a range of models from well-established, previously high ranking entries in the VQA2 competition to more recent, transformer-based architectures and report the insights we gained by inspecting them with our tool.

\subsection{Evaluated Models}
We chose a widely used \gls{vqa}-baseline BAN, two transformer-based architectures MDETR and MCAN, and MMNASNET.

\paragraph{BAN} \cite{ban} is a strong baseline using bilinear attention. It 
won third place in the VQA2 2018 challenge and was still in the top-10 entries in 2019. We use the 8-layer version.

\paragraph{MCAN} \cite{mcan} improves BAN with a co-attention feature fusion mechanism.

\paragraph{MMNASNET} \citet{mmnasnet} is a more recent state-of-the-art model constructed using neural architecture search. It is one of the top-10 entries of the VQA2 2020 challenge.

\paragraph{MDETR} \cite{kamath2021mdetr} is a state-of-the art transformer using more recent question \cite{liu2019roberta} and image embedding approaches \cite{carion2020end}.
MDETR achieves competitive accuracies on both GQA and CLEVR.

\subsection{Results and Lessons Learned}

Table \ref{tab:results} contains the aggregated results of all models, averaged across \added{the  development (sub-)splits of all datasets}.
\added{For details about the computation of each metric, see section \ref{sec:tool:metrics}.}
\begin{table*}[htb]
    \centering
    \resizebox{1.0\textwidth}{!}{ 
    \centering
    \begin{tabular}{|l|c|cc|cc|c|c|c|c|}
    \hline
    \multicolumn{10}{|c|}{\textbf{Average Results}}\\ \hline
    \multicolumn{1}{|c|}{{Model}} & \multicolumn{1}{c|}{Accuracy} & \multicolumn{2}{c|}{Modality Bias} & \multicolumn{2}{c|}{Robustness Image} & \multicolumn{1}{c|}{Robustness} & \multicolumn{1}{c|}{SEARs} & \multicolumn{1}{c|}{Uncer-} & \multicolumn{1}{c|}{Parameters} \\ \cline{3-6}
    & & Image & Quest. & Image & Feature & Question &  & tainty & \\ \hline
    MCAN & \textbf{41.30} & 4.83 &	2.57 & 99.98 & 81.09 & 63.14 & 58.45 & 31.77 & 201,723,191 \\
    MMNASNET & 40.80 & \textbf{4.67} & \textbf{2.51} & \textbf{99.99} & 79.90 & 65.18 & 59.65 & 31.75 & 211,166,871 \\
    BAN-8 & 38.62 & 5.21 & 3.31 &\textbf{99.99} & 82.22 & 70.09 & 53.53 & 66.10 & 112,167,258 \\
    MDETR & 38.82 & 4.77 & 2.60 & 36.33 & \textbf{90.10} & \textbf{100.00} & \textbf{100.00} & \textbf{24.58} & 185,847,022\\
    \hline
    \end{tabular}
    }
    \caption{Average results of our evaluated models across all development (sub-)splits.
    \added{We split the columns modality bias and robustness against image modifications into two sub-columns.}
    \added{These columns should be read as the modality bias  (lower values are better) measured for the image space and question space.}
    \added{Robustness (higher values are better) against image changes is divided into alterations on the image itself as well as modifications inside image feature space .}
    \added{All metrics are in range $[0,100]$. }} 
    \label{tab:results}
\end{table*}
Table \ref{tab:accuracy_detail} shows model accuracy per dataset.
Unsurprisingly, models performed best when evaluated on the development (sub-)split of the dataset they were originally trained on, and worse on datasets they were not trained on.
These performance drops are observable for all models, suggesting that \gls{vqa} models cannot yet generalize well across tasks. 
Low performance of current highly ranked \gls{vqa} models on new datasets can partially be attributed to their fixed answer spaces.
This implies the need for more research into systems that are able to generate answers instead of treating \gls{vqa} as a multiple-choice problem.
However, even changing distributions of the same dataset leads to a large performance drop, as we observe, for example, in GQA and its out-of-distribution variants, GQA-OOD-HEAD and GQA-OOD-TAIL.
By swapping the original GQA dataset for the GQA-OOD-TAIL distribution, MDETR accuracy decreased by more than $11,6\%$.
That out-of-distribution testing causes such high losses in accuracy indicates models are still relying on biases learned from the training dataset.
\begin{table*}[htb]
    \centering
    \resizebox{1.0\textwidth}{!}{ 
    \centering
    \begin{tabular}{|l|c|c|c|c|c|c|c|c|}
        \hline
        \textbf{Model} & \textbf{CLEVR} & \textbf{GQA} & \textbf{GQA-OOD-ALL} & \textbf{GQA-OOD-HEAD} & \textbf{GQA-OOD-TAIL} & \textbf{OK-VQA} & \textbf{TextVQA} & \textbf{VQA2} \\ \hline
        MCAN & 32.87 & 44.58 & 41.67 & 44.78 & 36.59 & 35.46 & 8.49 & \textbf{85.94} \\
        MMNASNET & 31.95 & 44.50 & 40.24 & 42.01 & 37.35 & 35.09 & 8.81 & \textbf{86.51} \\
        BAN-8 & 28.64 & 41.86 & 38.95 & 40.97 & 35.65 & 32.88 & 8.40 & \textbf{81.60} \\
        MDETR & 25.44 & \textbf{61.42} & 55.76 & 59.43 & 49.76 & 9.83 & 4.70 & 44.22 \\ \hline
    \end{tabular}
    }
    \caption{Accuracy across the \added{development (sub-)splits} of different datasets. Bold entries mark best accuracy per model and coincides in all cases with the dataset it was trained on.}
    \label{tab:accuracy_detail}
\end{table*}

All systems struggled to read text in images, in fact, the highest accuracy score on TextVQA was only $8.81\%$, achieved by MMNASNET.
This might be improved by extending existing \gls{vqa}-architectures with additional inputs, e.g. from optical character recognition, or adapting the training of currently used image feature extractors.

Applying noise in image space has almost no impact on models using bottom-up-topdown feature extraction \cite{butd}, in contrast to MDETR, the only model using an alternative approach.
In feature space, all models are similarly stable, which could imply that the feature extractor in MDETR could be made more robust by augmenting training with noisy images.

All models show highest modality bias and low accuracy on the CLEVR dataset. Given that no models were trained on synthetic images or questions involving such complex selection and spatial reasoning, this hints at the models not understanding either modality well.
Inspection using the filter view on modality biases provides more evidence of understanding problems here, showing that for example BAN-8 nearly always guesses \emph{yes} or \emph{no}, regardless of the question asked or the image given. 
In general, BAN-8 displays the highest modality bias, indicating more recent models have become better at jointly reasoning over image and text.

\Gls{sear} and question robustness metrics show that RoBERTa \cite{liu2019roberta} provides substantial robustness to question perturbation;
there were zero cases causing MDETR to change predictions, suggesting that context-aware embeddings should be a standard consideration for future \gls{vqa} models.

Our metrics show that state-of-the-art \gls{vqa} models are optimized for specific tasks or datasets, but fail to generalize even across other in-domain datasets.
In order to be successful in real-world applications, systems must demonstrate a variety of abilities, not merely good performance on a single-purpose test set.

\section{Conclusion}

Our proposed benchmarking tool is the first of its kind in the domain of \gls{vqa} and addresses the problems of current single-metric leaderboards in this domain.
It provides easy to use and fast comparison of integrated models on a global level.
The performance of each model is evaluated across multiple special-purpose as well as general-purpose datasets to test generalizability and capabilities.
Each model can be quantified by metrics such as accuracy, biases, robustness, and uncertainty, revealing strengths and weaknesses w.r.t to given tasks, i.e. measuring the properties models offer as well as their real-world robustness.
Exploration via filtering can be used to identify suspicious behaviour down to single data sample level.
Through this, our tool provides deeper insights into the strengths and weaknesses of each model across tasks and metrics and how architectural choices can affect behavior, encouraging researchers to develop \gls{vqa} systems with rich sets of abilities that stand up to real-world environments.
The open-source tool itself can be installed as a package and extended with new models, datasets and metrics using our python API.

In the future, we plan to extend this tool with new datasets as they are released.
Moreover, we are looking for more metrics for model evaluation as well as more detailed dataset analysis, e.g. answer space overlap. 
Last but not least, interactivity could be extended towards live model feedback, allowing to change inputs, e.g. the image noise level, and observe model outputs at runtime.

\section{Acknowledgement}
This research was funded by the Cluster of Excellence EXC 2075 "Data-Integrated Simulation Science" at the University of Stuttgart.

\bibliography{anthology,custom}
\bibliographystyle{acl_natbib}

\appendix

\section{Appendix}
\label{sec:appendix}

\begin{figure}[h]
    \centering
    \includegraphics[width=0.49\textwidth]{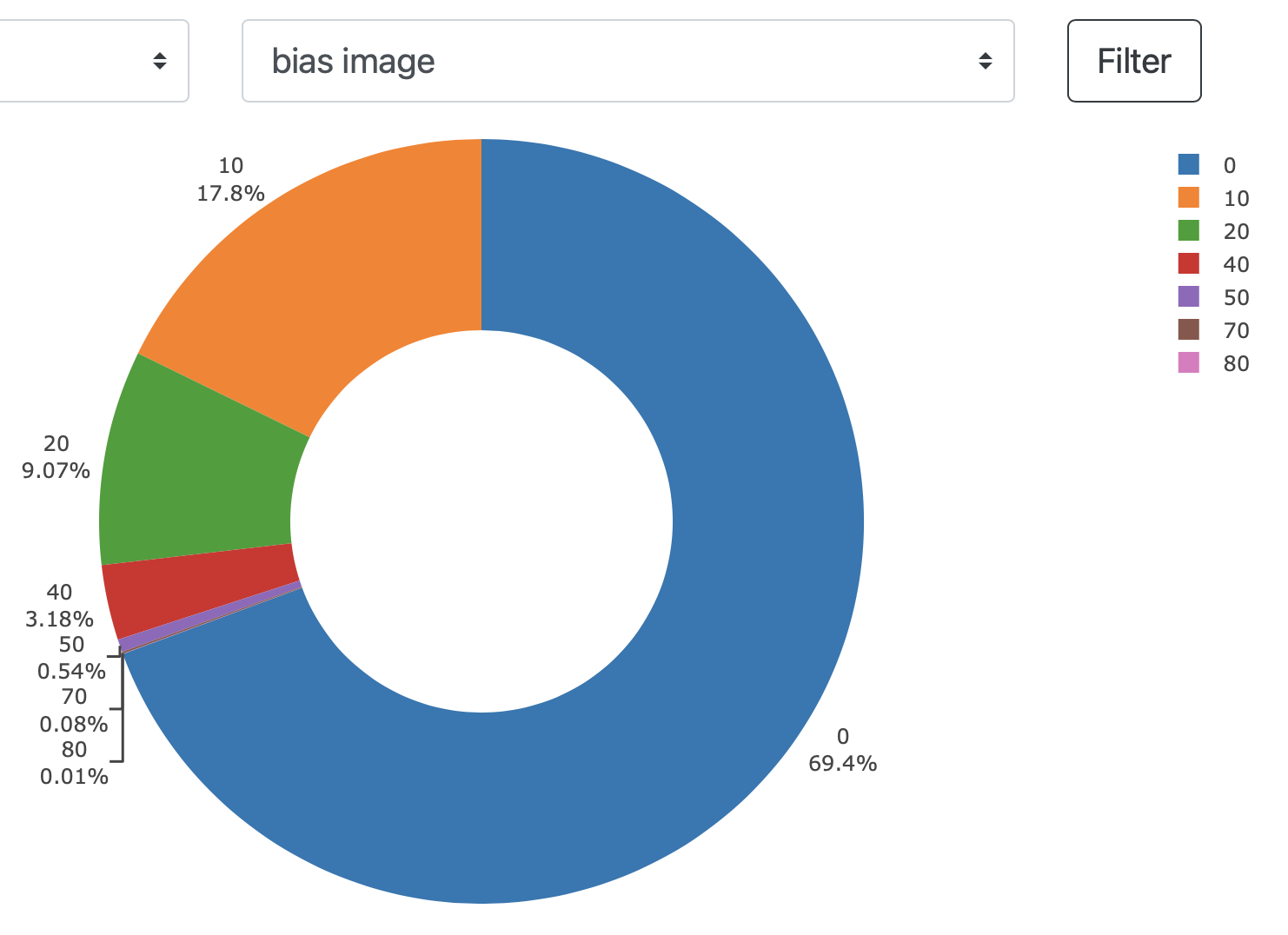}
    \caption{Zoom into Figure \ref{fig:metric_view}.}
    \label{fig:metric_view_appendix}
\end{figure}

\onecolumn
\begin{figure*}[htb]
    \centering
    \includegraphics[width=\textwidth]{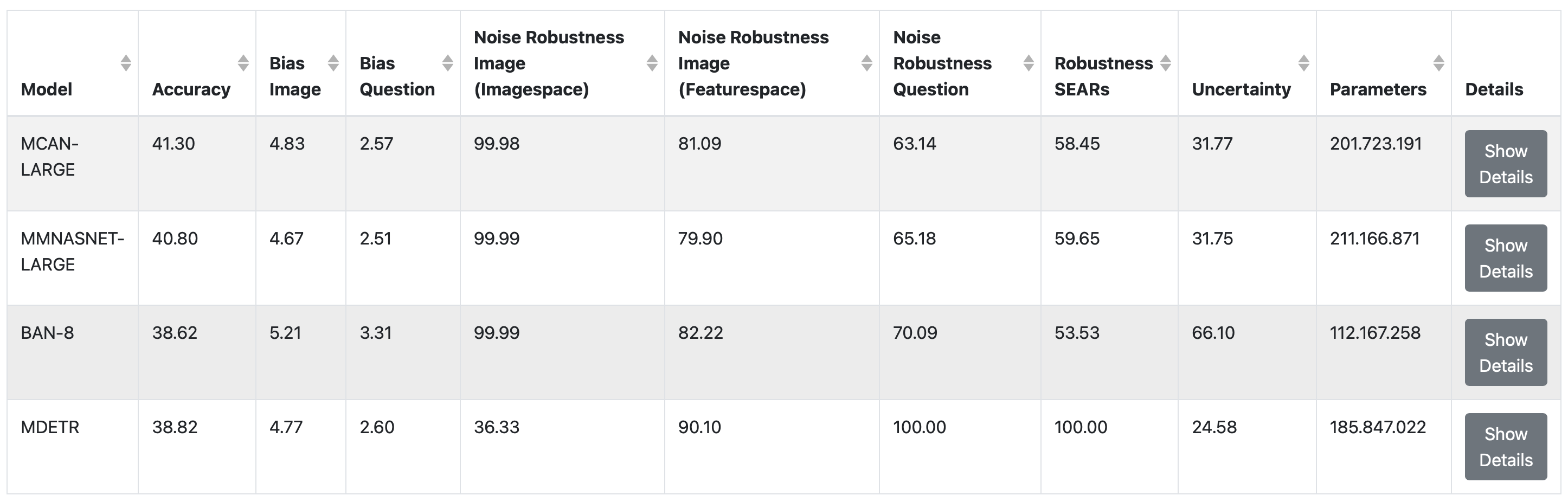}
    \caption{Figure \ref{fig:overview} in full size.}
    \label{fig:overview_appendix}
\end{figure*}
\begin{figure*}[htb]
    \centering
    \includegraphics[width=\textwidth]{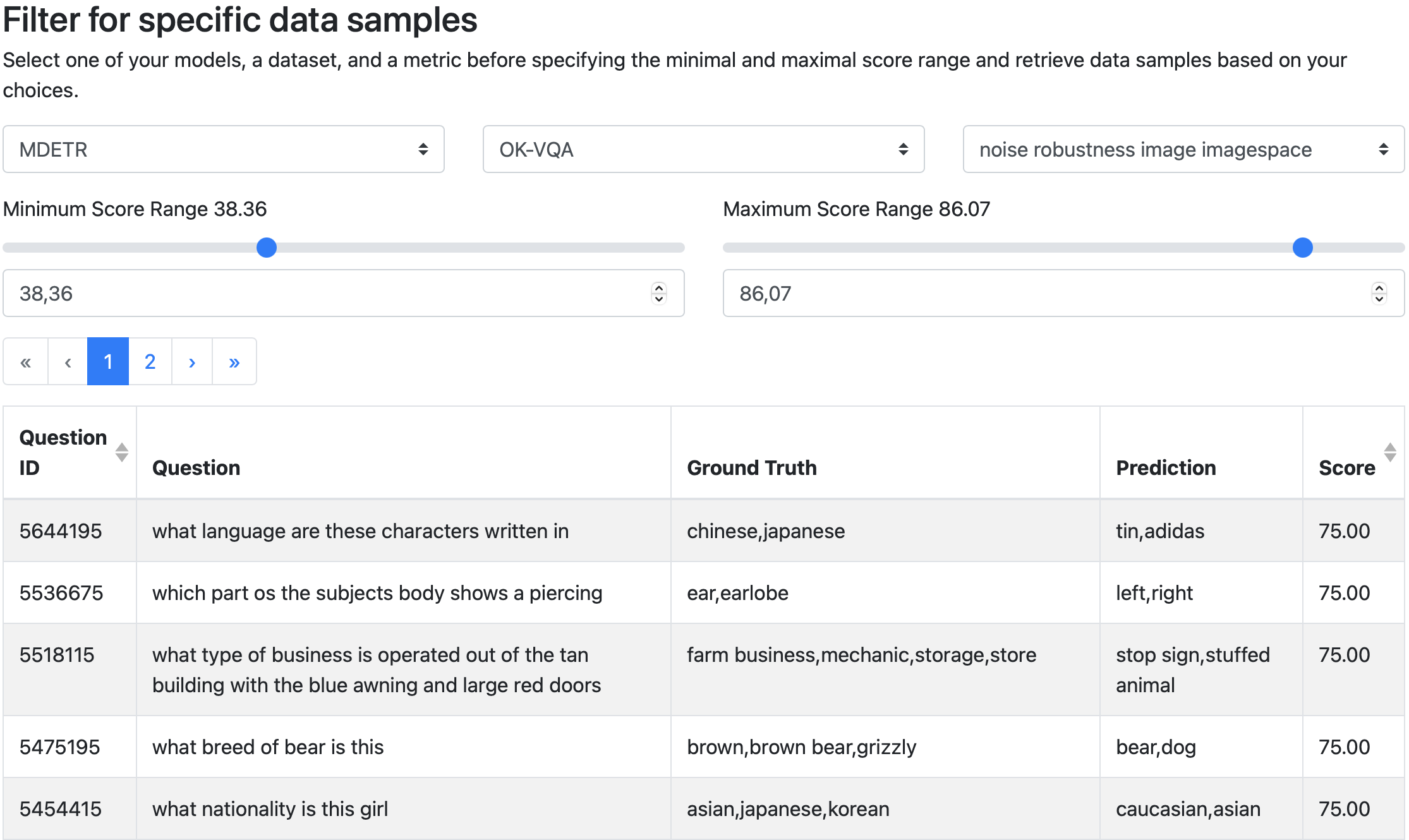}
    \caption{Figure \ref{fig:filter_view} in full size.}
    \label{fig:filter_view_appendix}
\end{figure*}
\begin{figure*}[htb]
    \centering
    \includegraphics[width=\textwidth]{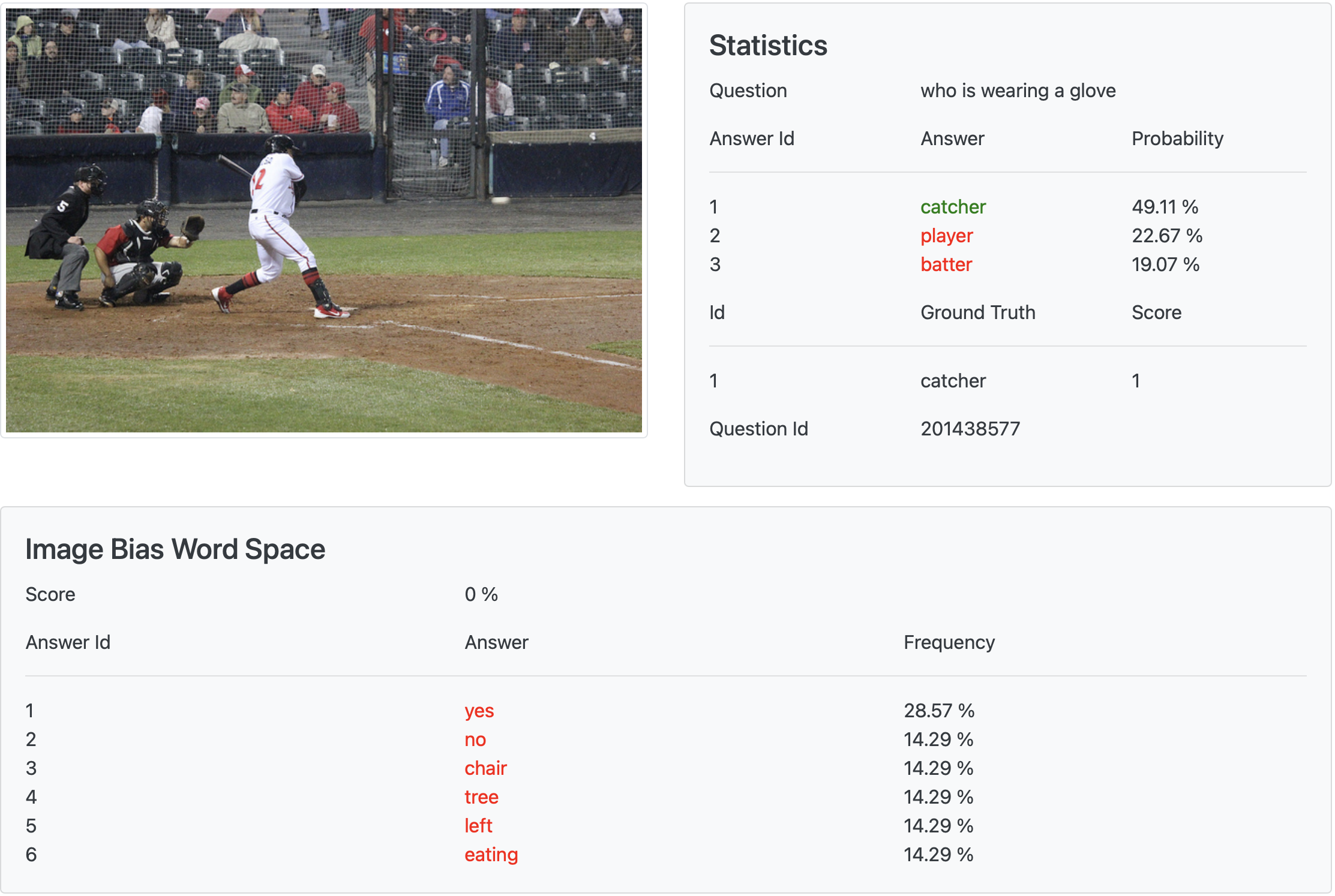}
    \caption{Figure \ref{fig:sample_view} in full size.}
    \label{fig:sample_view_appendix}
\end{figure*}

\end{document}